\documentclass{article}

\usepackage[preprint,nonatbib]{neurips_2024}

\usepackage{graphicx}
\usepackage{booktabs}
\usepackage{multirow}
\usepackage{pifont}
\usepackage{array}

\usepackage[utf8]{inputenc} 
\usepackage[T1]{fontenc}    
\usepackage{hyperref}       
\usepackage{url}            
\usepackage{booktabs}       
\usepackage{amsfonts}       
\usepackage{nicefrac}       
\usepackage{microtype}      
\usepackage{xcolor}         
\usepackage{url}
\usepackage{xspace}
\usepackage{fontawesome5}

\usepackage{dirtree}

\usepackage{graphicx}
\graphicspath{{figs}{figs/refers}}
\usepackage{subcaption}

\usepackage{booktabs}
\usepackage{multirow}
\PassOptionsToPackage{table,xcdraw,svgnames,dvipsnames}{xcolor}
\usepackage{floatrow}
\newfloatcommand{capbtabbox}{table}[][\FBwidth]
\usepackage{wrapfig}

\usepackage{algorithm}
\usepackage{algorithmicx}
\usepackage[noend]{algpseudocode}
\algrenewcommand\alglinenumber[1]{\tiny #1.}
\usepackage{adjustbox}

\usepackage{stmaryrd}
\usepackage{svg}
\definecolor{indianred}{rgb}{0.8, 0.36, 0.36}
\definecolor{bleudefrance}{rgb}{0.19, 0.55, 0.91}
\definecolor{forestgreen}{rgb}{0.0, 0.5, 0.0}
\definecolor{ashgrey}{rgb}{0.7, 0.75, 0.71}
\definecolor{darkorange}{rgb}{1.0, 0.55, 0.0}
\definecolor{darkorchid}{rgb}{0.6, 0.2, 0.8}

\newcommand{\icoyes}{\textcolor{forestgreen}{\faIcon{check-circle}}\xspace}
\newcommand{\icohalf}{\textcolor{darkorange}{\faIcon{check-circle}}\xspace}
\newcommand{\icono}{\textcolor{ashgrey}{\faIcon{times-circle}}\xspace}

\newcommand{\Nimages}{29,614\space}
\newcommand{\Nref}{52,472\space}
\newcommand{\Nvqa}{123,221\space}
\newcommand{\papernameAbbrev}{VRSBench}

\newcommand{\gitpage}{https://github.com/lx709/VRSBench}
\newcommand{\prjpage}{https://vrsbench.github.io}
\newcommand{\huggingpage}{https://huggingface.co/datasets/xiang709/VRSBench}

\title{VRSBench: A Versatile Vision-Language Benchmark Dataset for Remote Sensing Image Understanding}

\author{%
Xiang Li \quad Jian Ding \quad Mohamed Elhoseiny \\
King Abdullah University of Science and Technology \\
\texttt{\{xiang.li.1,jian.ding,mohamed.elhoseiny\}@kaust.edu.sa}\\
}

\begin{document}

\maketitle

\begin{abstract}
We introduce a new benchmark designed to advance the development of general-purpose, large-scale vision-language models for remote sensing images. Although several vision-language datasets in remote sensing have been proposed to pursue this goal, existing datasets are typically tailored to single tasks, lack detailed object information, or suffer from inadequate quality control. Exploring these improvement opportunities,  we present a \textbf{V}ersatile vision-language \textbf{Bench}mark for \textbf{R}emote \textbf{S}ensing image understanding, termed \textbf{\papernameAbbrev}. This benchmark comprises {\Nimages} images, with \Nimages human-verified detailed captions, \Nref object references, and \Nvqa question-answer pairs. It facilitates the training and evaluation of vision-language models across a broad spectrum of remote sensing image understanding tasks. We further evaluated state-of-the-art models on this benchmark for three vision-language tasks: image captioning, visual grounding, and visual question answering. Our work aims to significantly contribute to the development of advanced vision-language models in the field of remote sensing. The data and code can be accessed at \url{\prjpage}.
\end{abstract}

\begin{figure}[!htbp]
    \centering
    \includegraphics[width=\textwidth]{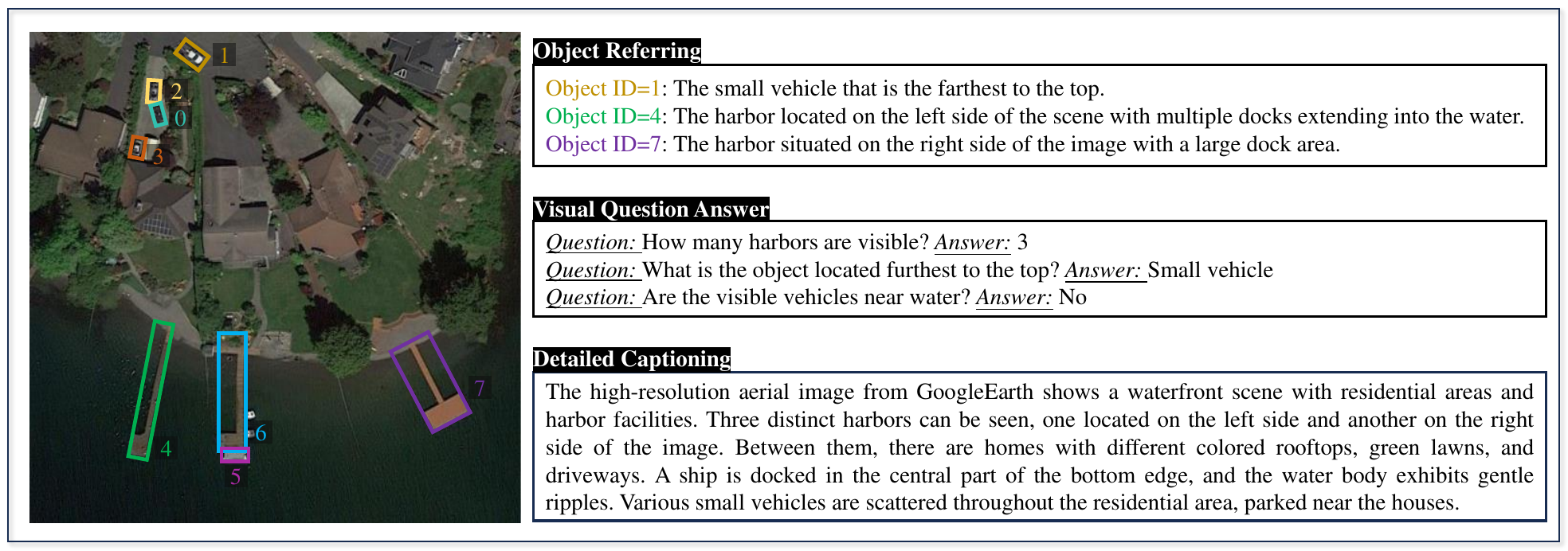}
    \caption{Examples of an image and corresponding annotations in {\papernameAbbrev} dataset. Our annotations include object referring, visual question answering, and detailed captions.}
    \label{fig_emp}
\end{figure}

\section{Introduction}\label{sc_intro}
Remote sensing models seek to understand the Earth's surface using imagery captured from overhead, offering a unique perspective of our physical world. This technique is instrumental in various applications, such as land use mapping, urban planning, precision agriculture, disaster management, etc. 
In the past few years, the success of large vision-language models (LVLMs)\cite{chatgpt,gemini,liu2024visual,zhu2024minigpt,bai2023qwen} in natural scenes has inspired a trend of applying LVLMs to remote sensing~\cite{li2024vision}. Recent efforts have explored LVLMs for various remote sensing image understanding tasks, including scene classification~\cite{liu2024remoteclip,li2023rs}, image captioning~\cite{shi2017can,lu2017exploring,zhang2017natural,zhang2019description,li2020multi,wang2020word,li2020truncation,zhao2021high,zia2022transforming}, visual grounding~\cite{sun2022visual,zhan2023rsvg,liu2024rotated}, visual question answering (VQA)~\cite{lobry2020rsvqa,zheng2021mutual,chappuis2022language,al2022open,bazi2022bi,yuan2022easy}, and general-purpose models~\cite{hu2023rsgpt,kuckreja2024geochat}, etc.

However, directly applying LVLMs to remote sensing images presents challenges. LVLMs are typically trained on \textit{internet data}, which differs significantly from remote sensing data. Remote sensing images often feature very small objects (sometimes only 10 pixels) and require complex spatial reasoning from an overhead view. Building effective LVLMs for remote sensing requires large-scale, high-quality datasets tailored to this field. Recent works\cite{hu2023rsgpt,kuckreja2024geochat,zhang2024earthgpt} have attempted to train LVLMs with a combination of existing text-enriched remote sensing data, achieving reasonable performance. However, further improvements are limited by the current vision-language datasets in remote sensing, which have the following limitations:

\textbf{(i)} Existing vision-language datasets primarily cater to single image perception tasks, e.g., image captioning. Recent works explore integrating multiple datasets to accommodate a wider array of tasks~\cite{kuckreja2024geochat,zhang2024earthgpt}. Such integration, while crucial, introduces challenges including inconsistent data annotations, variations in data quality, and the complexity of merging different data formats and sources, all of which can hinder model performance and scalability. 

\textbf{(ii)} Most commonly used remote sensing image caption datasets, such as UCM-Captions~\cite{qu2016deep} and RSICD~\cite{lu2017exploring}, provide only brief descriptions, lacking detailed object information. Recent work RSGPT~\cite{hu2023rsgpt} provides high-quality, human-generated detailed image captions; however, the dataset comprises only 2,585 image-text pairs. This limited scope restricts its potential for training robust vision-language models in remote sensing applications. Although recent works, such as RS5M~\cite{zhang2023rs5m} and RemoteClip~\cite{liu2024remoteclip}, introduced large-scale remote sensing image-text pair datasets, these annotations are automatically generated by image caption models and lack human verification. Given the current limitations of automatic captioning technology, such image-text data often suffer from accuracy issues and a lack of quality control.

\textbf{(iii)} Most existing remote sensing visual grounding datasets are designed under simplistic scenarios where the referring objects typically stand alone within their category. For instance, in the widely used DIOR-RSVG~\cite{zhan2023rsvg} datasets, a large portion of objects are unique within the categories, which leads to 38.36\% of objects being easily distinguished by the object category alone. Finally, the majority of current VQA datasets in remote sensing employ automated methods for generating question-answer pairs. These automatically generated pairs often encompass a limited variety of unique questions and answers, which may not be sufficiently diverse to facilitate open-ended question-answering in real-world applications.

In this study, to address these limitations, we introduce a novel versatile benchmark for vision-language understanding of remote sensing images.
{\papernameAbbrev} comprises \Nimages images, each enriched with human-verified detailed captions, complex object referring, and question-answer pairs, check Table \ref{tab_dataset} for a detailed comparison with existing datasets. This dataset facilitates the training and evaluation of vision-language models across a spectrum of remote sensing image understanding tasks. Fig.~\ref{fig_emp} gives an example of a selected image and associated annotations. 

The key contributions of our work are summarized as follows:

\begin{itemize}
    \item We introduce a new semi-automatic vision-language data collection pipeline which includes four key steps: object attributes extraction, prompt engineering, GPT-4 inference, and human verification. This pipeline enables a fast collection of large-scale datasets with human-level annotation quality.
    \item Based on the semi-automatic data collection pipeline, we collect {\papernameAbbrev} dataset that provides detailed image captioning, visual grounding, and visual question-answer labels in a unified dataset, and therefore, enables a comprehensive evaluation of multiple vision-languages capabilities based on this dataset.
    \item {\papernameAbbrev} provides large-scale human-verified annotations that feature several advantages: 1) it provides a large-scale collection of human-verified, high-quality captions rich in object details; 2) it offers more realistic object refers in which each referring sentence unambiguously identifies an object among multiple similar ones within the same category; 3) it features a diverse collection of open-ended question-answer pairs in natural language. 
    \item We develop three benchmarks based on our {\papernameAbbrev} dataset, including detailed image caption, visual grounding, and visual question answering, and evaluate the performance of several state-of-the-art LVLMs.
\end{itemize}

\begin{table}[!htbp]
    \centering
    \resizebox{0.98\textwidth}{!}{
    \begin{tabular}{l|c|cccccccc}
        \toprule
        \multirow{2}{*}{Dataset} & \multirow{2}{*}{Year} & \multirow{2}{*}{\#Image} & \multicolumn{2}{c}{Caption} & \multicolumn{2}{c}{Grounding} & \multicolumn{2}{c}{VQA} & Human \\
        \cline{4-5} \cline{6-7} \cline{8-9}
         &  &  & \#Captions & Details & \#Refers & OBB & \#VQAs & Open-ended &  \\
        \hline
        UCM-Captions~\cite{qu2016deep} & 2016 & 2,100 & 10,500 (12) & \icono & 0 & \icono & 0 & - & \icoyes \\
        RSICD \cite{lu2017exploring} & 2017 & 10,921 & 54605 (12) & \icono & 0 & \icono & 0 & - & \icoyes \\
        RS5M \cite{zhang2023rs5m} & 2023 & 5M & 5M (49) & \icoyes & 0 & \icono & 0 & - & \icono \\
        RSICap~\cite{hu2023rsgpt} & 2023 & 2,585 & 2,585 (60) & \icoyes & 0 & \icono & 0 & - & \icoyes\\
        \hline
        RSVG~\cite{sun2022visual} & 2022 & 4,239 & 0 & \icono & 7,933 & \icono & 0 & - & \icoyes \\
        DIOR-RSVG \cite{zhan2023rsvg} & 2023 & 17,402 & 0 & \icono & 38,320 & \icono & 0 & - & \icoyes \\
        RRSIS-D \cite{liu2024rotated} & 2024 & 17,402 & 0 & \icono & 17,402 & \icoyes & 0 & - & \icoyes \\
        \hline
        RSVQA-HR \cite{lobry2020rsvqa} & 2020 & 10,659 & 0 & \icono & 0 & \icono & 1,066,316 & \icono & \icono \\
        RSIVQA \cite{zheng2021mutual} & 2021 & 37,264 & 0 & \icono & 0 & \icono & 111,134 & \icohalf & \icohalf \\
        VQA-TextRS \cite{al2022open} & 2022 & 2,144 & 0 & \icono & 0 & \icono & 6,245 & \icoyes & \icoyes \\
        RSIEval \cite{hu2023rsgpt} & 2023 & 100 & 0 & \icono & 0 & \icono & 933 & \icoyes & \icoyes \\
        \hline
        {\papernameAbbrev} & 2024 &  \Nimages & \Nimages (52) & \icoyes & \Nref & \icoyes & \Nvqa & \icoyes & \icoyes \\
        \bottomrule
    \end{tabular}
    }
    \caption{Comparison between existing remote sensing vision-language datasets and our {\papernameAbbrev} dataset. Values in parentheses in the Caption column indicate the average number of words in captions. OBB denotes orientated bounding box. A small portion of question-answer pairs in RSIVQA are annotated by human annotators.}
    \label{tab_dataset}
\end{table}

\section{Pipeline}\label{sc_{\papernameAbbrev}}
To construct our {\papernameAbbrev} dataset, we employed multiple data engineering steps, including attribute extraction, prompting engineering, GPT-4 inference~\cite{gpt4}, and human verification. These processes are meticulously designed to ensure the integrity and utility of the dataset for remote sensing applications.

\begin{figure}[!htbp]
    \centering
    \includegraphics[width=\textwidth]{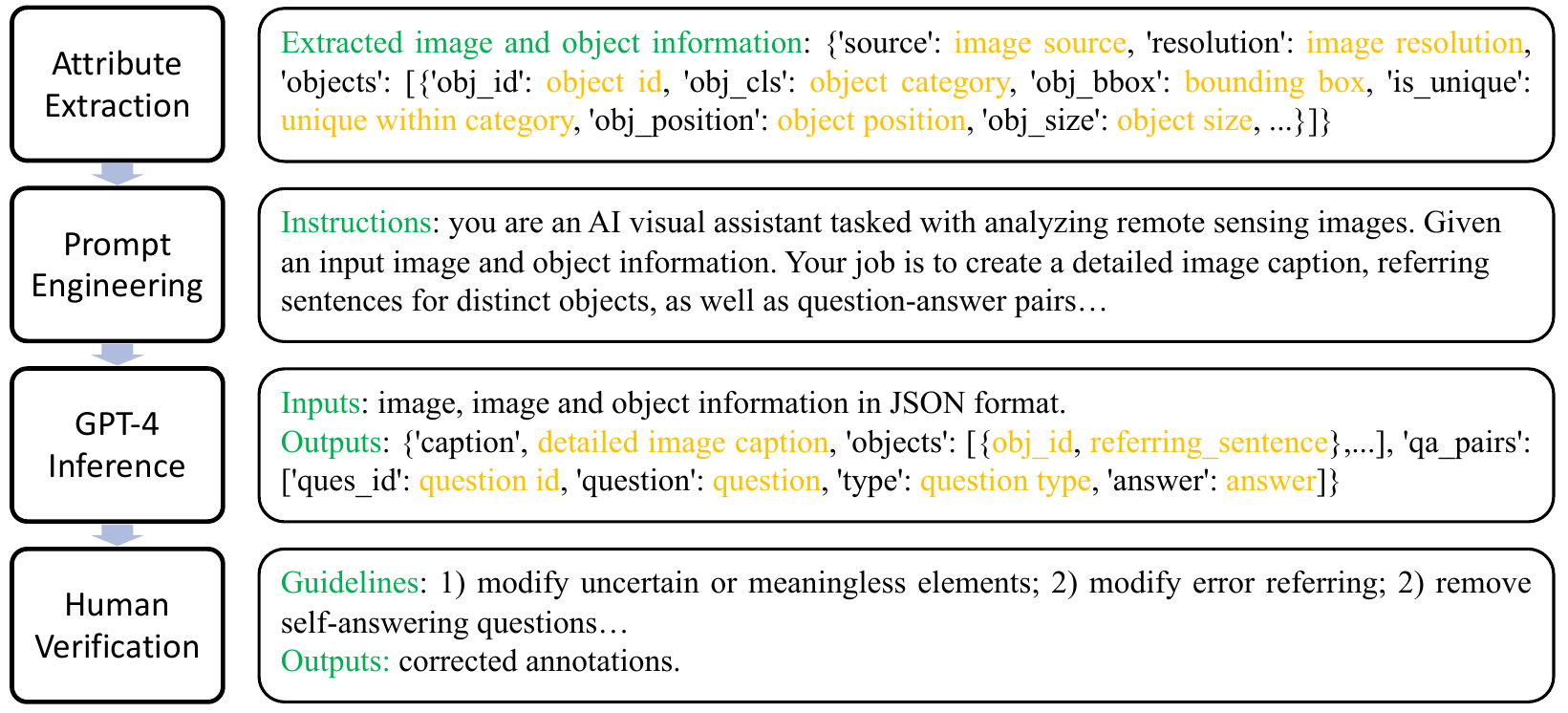}
    \caption{Dataset creation pipeline. We generate object information from detection labels and use carefully designed instructions to prompt GPT-4 to generate annotations from input images along with object information. All annotations are verified by human annotators.}
    \label{fig_pipeline}
\end{figure}

\subsection{Attribute Extraction}
Initially, we extract image information, including the source and resolution, as well as object information—such as the object category, bounding box, color, position (absolute and relative), and size (absolute and relative)—from existing object detection datasets. 
We also determine whether an object is unique within its category, which is important for crafting accurate reference sentences. 

In this study, we utilize two prominent open-access object detection datasets, DOTA-v2~\cite{ding2021object} and DIOR~\cite{li2020object}, to develop our {\papernameAbbrev} dataset. Due to the unavailability of test labels for DOTA-v2, we incorporate only its training and validation sets. We divide each image into patches measuring 512 × 512 pixels. Notably, each image patch from DOTA-v2 contains, on average, 14.2 instances, while each patch from the DIOR dataset averages only 3.3 instances. This higher instance density in DOTA-v2 offers a more challenging and diverse training environment compared to existing remote sensing visual grounding datasets, such as DIOR-RSVG~\cite{zhan2023rsvg} and RRSIS-D~\cite{yuan2024rrsis}, that are typically sourced from the DIOR dataset. Moreover, the visual grounding task predominantly involves identifying horizontal bounding boxes (HBB) based on referential descriptions. By constructing our dataset upon the framework of DOTA-v2, {\papernameAbbrev} facilitates the grounding of objects with orientated bounding boxes (OBB), thereby extending the capabilities of traditional visual grounding methods.

\subsection{Prompt Engineering}
We carefully design the following instructions to prompt GPT-4V to create detailed image captions, object referring, and question-answer pairs. Detailed instructions for each task are provided in the supplementary. 

``You are an AI visual assistant tasked with analyzing remote sensing images. For each image, you receive image meta information and a list of objects in the format of {...}. Your job is to create a detailed image caption and referring sentences for 1-5 distinct objects, if multiple are present, as well as 3-10 question-answer pairs. Each referring sentence should unambiguously refer to one object. Finally, you need to return a JSON file in the format: \{caption: detailed image caption, objects: [{obj\_id: object id, ref: referring sentence},...], qa\_pairs: [ques\_id: question id, question: question, type: question type, answer: answer]\}. Do not return any notes after the JSON.''

\subsection{GPT-4V Inference}
Given input prompts, we call OpenAI API\footnote{https://platform.openai.com/docs/api-reference} to automatically generate annotations. We iteratively refine our instructional prompts to generate annotations, meticulously enhancing these instructions to ensure the quality of the annotations.
In the responses generated by GPT-4V, undesirable terms, such as ``not provide'', ``not specified'', and ``unknown'', may be present. Should any of the specified excluding phrases appear in GPT-4V’s output, the procedure requires that GPT-4V be recursively invoked to regenerate responses until the output is free of any excluding phrases. This iterative process is attempted a maximum of five times, after which the final response is utilized for generating annotations. Ultimately, any caption sentences, object-referring sentences, or question-answer pairs containing these excluding phrases are excised from final annotations.

\subsection{Human Verification}
With our carefully designed prompts, most of the annotations generated by GPT-4V are accurate. Nevertheless, a significant number of outputs remain suboptimal. This shortfall is likely attributable to the model's limited exposure to remote sensing imagery, which impedes its capacity to interpret complex structures within these images. Additionally, it is important to note that even advanced language models, such as the GPT-4V system, exhibit a degree of hallucinatory outputs~\cite{liu2024survey}. 

To improve the quality of the dataset, we engage human annotators to validate each annotation generated by GPT-4V. This validation process incorporates domain experts to guarantee that annotators have a comprehensive understanding of the assigned tasks. Initially, domain experts establish detailed guidelines, which include directives such as: 1) eliminate any uncertain or irrelevant elements; 2) ensure each referring sentence unambiguously identifies the intended object; 3) exclude questions that inherently contain their answers. More details about the annotation guidelines can be found in the supplementary. The verification of each image requires approximately 120 seconds, culminating in a total of 1,004 hours devoted to human verification. Each image verification costs around 0.21 USD and leads to a total cost of 6,200 USD for human verification. To enhance the quality of our dataset, we have instituted a secondary validation phase involving a meticulous re-evaluation of 2,000 images. This step is designed to uncover prevalent annotation discrepancies and to refine the annotators' understanding of the task requirements.

\section{{\papernameAbbrev} Dataset}
\subsection{Dataset Overview}
Our {\papernameAbbrev} dataset contains \Nimages remote sensing images, with high-quality human-verified annotations. It comprises \Nimages caption sentences, \Nref referring sentences, and \Nvqa question-answer pairs. Each image is of $512 \times 512$ pixels. Details of each type of annotation are given below. Note that original object detection labels and object attributes are also provided in our annotations.

\subsection{Detailed Caption}
\begin{figure}[!htbp]
    \centering
    \begin{subfigure}[b]{0.26\textwidth}
        \includegraphics[width=\textwidth]{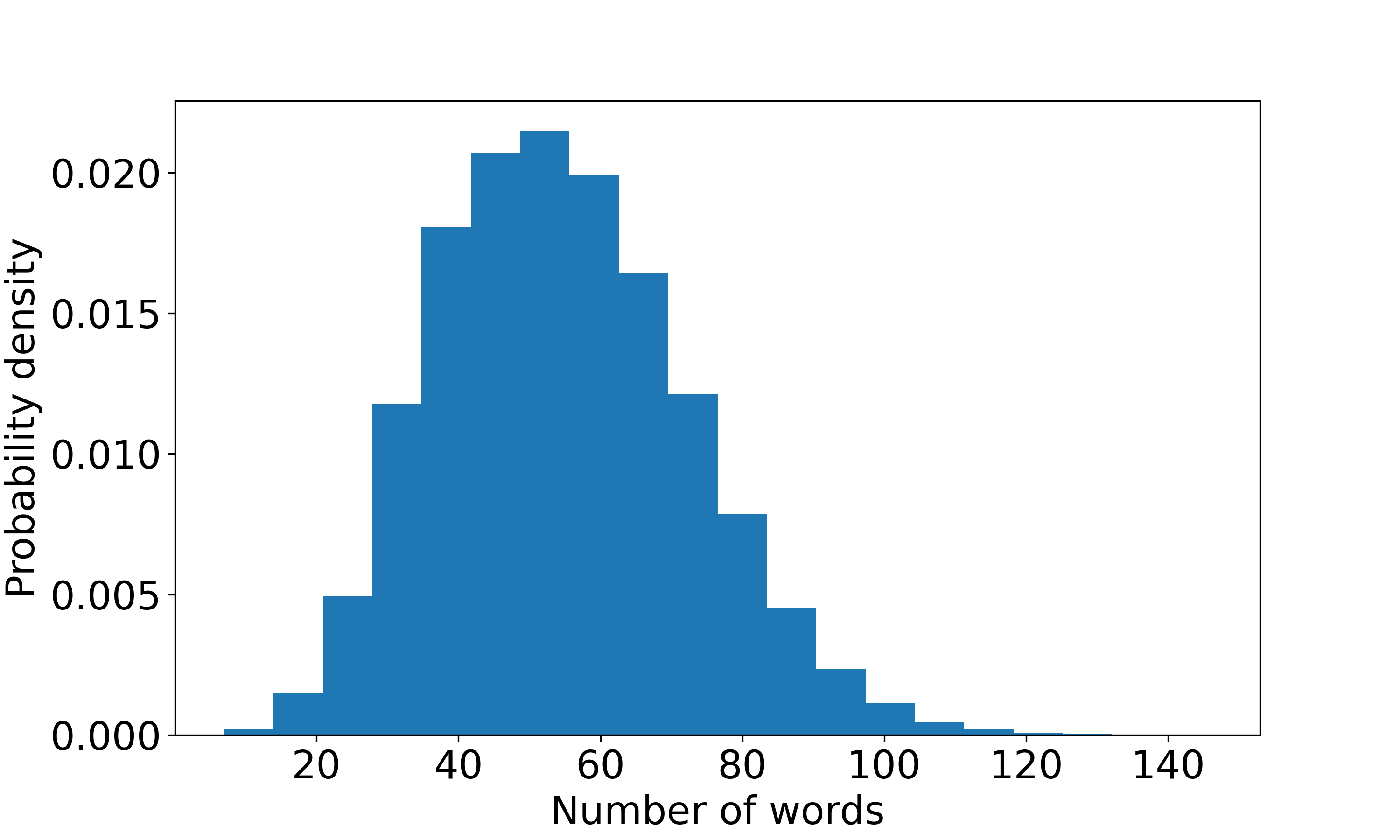}
        \caption{PDF of word count}
    \end{subfigure}
    \hfill
    \begin{subfigure}[b]{0.26\textwidth}
        \includegraphics[width=\textwidth]{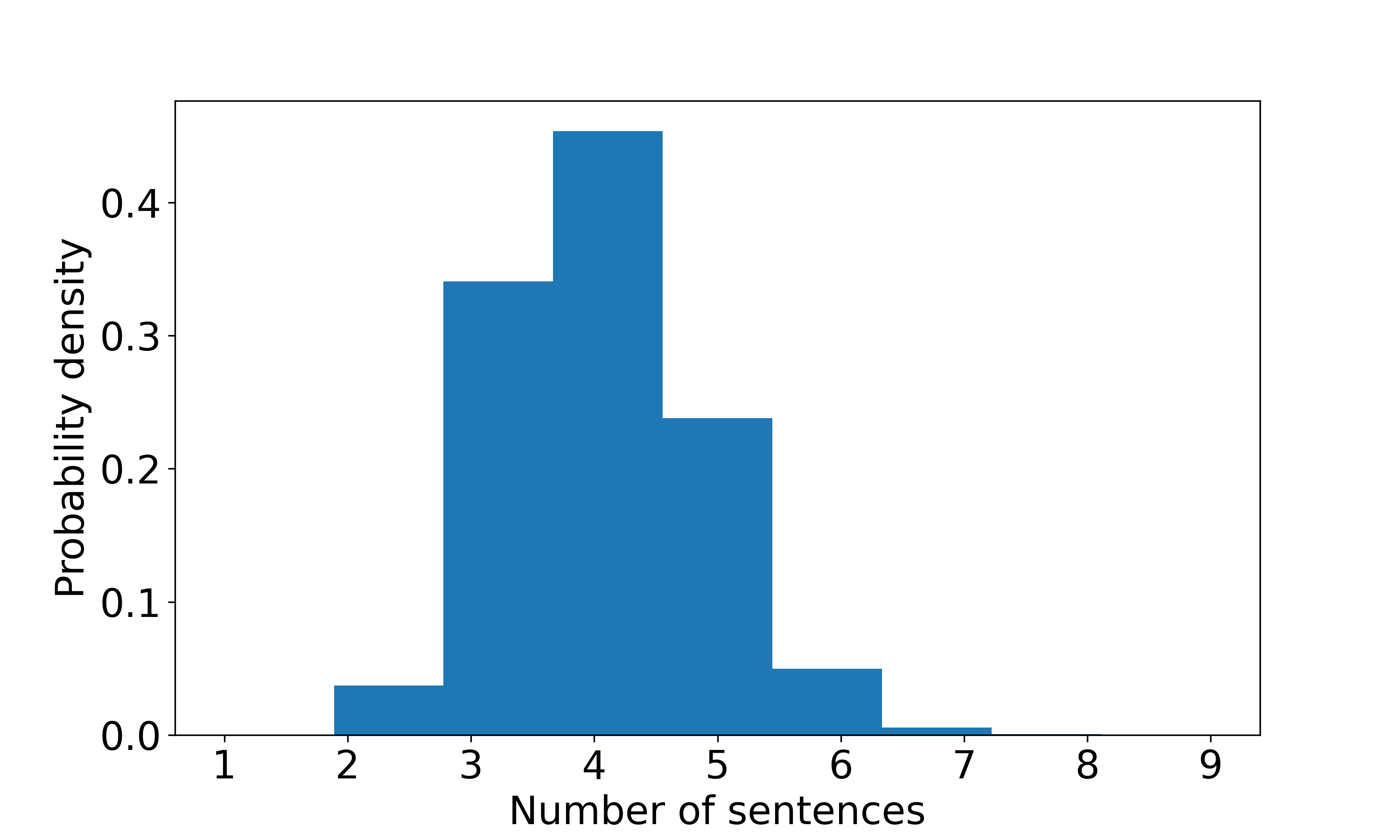}
        \caption{PDF of sentence number}
    \end{subfigure}
    \hfill
    \begin{subfigure}[b]{0.4\textwidth}
        \centering
        \resizebox{\columnwidth}{!}{%
        \begin{tabular}{cc}
        \hline
        \#images & \Nimages \\
        \#vocabulary size & 9,588 \\
        \#total words & 1,526,338 \\
        \#caption sentences & 114,366 \\
        Avg. \#sentences in caption  & 4 \\
        Avg. caption length & 52 \\
        \hline
        \end{tabular}
        }
        \caption{Statistics of {\papernameAbbrev} captions.}
    \end{subfigure}
    \caption{Statistics of the {\papernameAbbrev} caption dataset. (a) Probability density function (PDF) of caption length. (b) PDF of the sentence number. (c) Summative statistics.}
    \label{fig_stat_cap}
\end{figure}

{\papernameAbbrev} captions provide comprehensive descriptions that encompass both abstract image attributes and detailed object-specific information. Each caption initiates with a general overview of the image, subsequently delving into explicit and precise details present within the image. Attributes of the image include the source, resolution, color or panchromatic distinction, and the type of scene depicted. Conversely, object attributes cover object quantity, color, shape, size, and spatial positioning of each object, encompassing both its absolute location within the image and its relative positioning in relation to other objects. Descriptions are confined to manifest features, eschewing any elements that are uncertain or ambiguous. Additionally, captions may incorporate other visually discernible objects not supplied by the source object detection datasets, such as buildings, houses, roads, and trees, if these elements are clear and unambiguous. Each caption typically comprises 3-7 sentences, with an average length of 54 words. A summary of these caption statistics is detailed in Fig. \ref{fig_stat_cap}.  

\subsection{Object Referring}
\begin{figure}[!htbp]
    \centering
    \includegraphics[width=\textwidth]{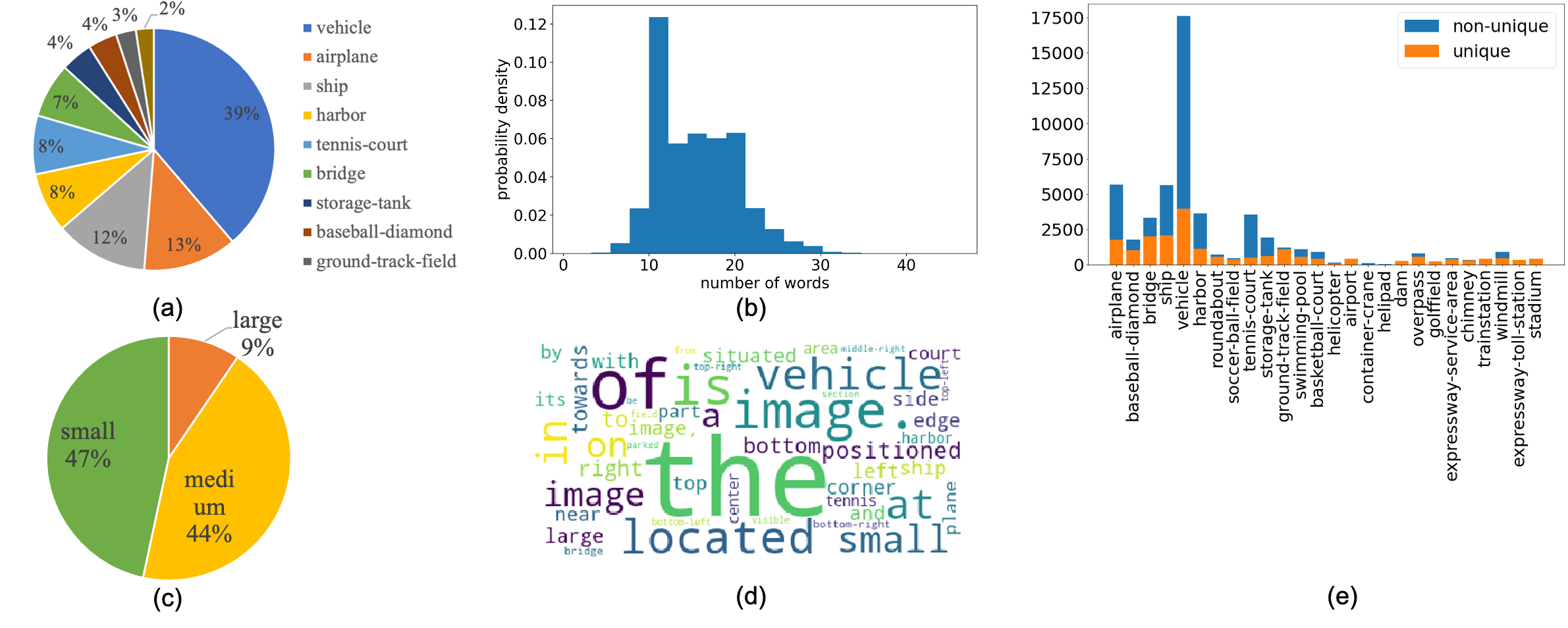}
    \caption{Statistics of object referring sentences of {\papernameAbbrev} dataset. (a) Distribution of the 10 most frequent object categories. (b) Distribution of the word length of referring sentences. (c) Distribution of object size. (d)Word cloud of the top 50 words in referring sentences. (e) Distribution of unique/non-unique objects in each category.}
    \label{fig_stat_ref}
\end{figure}

In {\papernameAbbrev}, each image is analyzed to identify 1-5 distinct objects, and referring sentences are provided for each. These sentences are carefully crafted such that each can independently and unambiguously identify an object without reliance on other sentences. We utilize distinctive features to clearly differentiate the referred objects from others within the image. These features span a variety of object attributes including color, shape, position, size, relative position, and relative size, among others. Note that the original DOTA-v2 and DIOR datasets contain 18 and 20 object categories respectively, which are merged into 26 object categories in our dataset. Please check the supplementary for category merging details. Figure~\ref{fig_stat_ref} provides a summary of referring sentences of our {\papernameAbbrev} dataset.

\subsection{Visual Question Answering}
\begin{figure}[!htbp]
    \centering
    \includegraphics[width=0.95\textwidth]{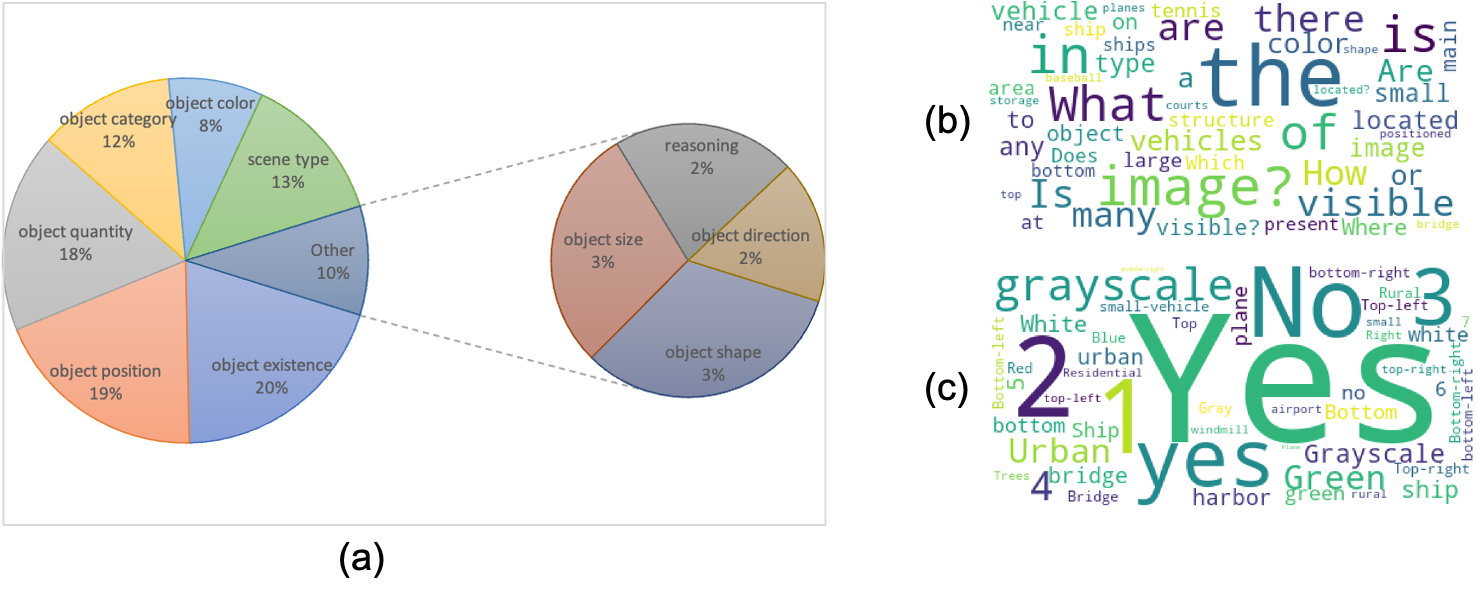}
    
    \caption{Statistics of question-answer pairs in {\papernameAbbrev}. (a) Distribution of question types. (b) Word cloud of top 50 most frequent words in questions. (c) Word cloud of top 50 most frequent words in answers.}
    \label{fig_stat_qa}
\end{figure}

Based on all visible elements and object information, we provide 3-10 question-answer pairs about diverse types, including object category, object existence, object quantity, object color, object size, object position, direction, scene characteristics, and complex reasoning, and provide an answer for each question. Instead of only focusing on objects from source detection datasets, we also ask questions about objects that are not provided, such as houses, roads, and trees if they are obvious and non-ambiguous. When collecting annotations, we ensure each question has a definite answer without any ambiguity, and answer each question using a single word or phrase. We show the statistics of question-answer pairs in Figure~\ref{fig_stat_qa}.

\section{Benchmark Evaluation}
\subsection{Benchmark Overview}
Based on {\papernameAbbrev}, we construct three distinct tasks for advancing remote sensing image understanding:
\begin{itemize}
    \item {\papernameAbbrev}-Cap: This challenge requires the prediction of a comprehensive description for a given remote sensing image, encapsulating intricate object details and contextual relevance.
    \item {\papernameAbbrev}-Ref: The task involves identifying and localizing specific objects from a given remote sensing image based on textual descriptions.
    \item {\papernameAbbrev}-VQA: This task aims to answer questions related to visual content in a given remote sensing image.
\end{itemize}

To facilitate benchmark evaluation, we partition our {\papernameAbbrev} dataset into two distinct, non-overlapping splits designated for model training and evaluation. We split the datasets according to official splits of DOTA~\cite{ding2021object} and DIOR~\cite{li2020object} datasets, where their training images are used to build the training set of {\papernameAbbrev} and their validation sets are used as the test set. Table \ref{tab_split} delineates the statistics of two splits.

\begin{wraptable}{r}{0.35\textwidth}
    \centering
    \caption{{\papernameAbbrev} data split.}
    \label{tab_split}
    \vspace{-0.2cm}
    \begin{tabular}{l c c}
        \hline
        & train & test \\
        \hline
        \#Images & 20,264 & 9,350  \\
        \#Captions & 20,264 & 9,350  \\
        \#Refers & 36,313 & 16,159  \\
        \#VQAs & 85,813 & 37,408 \\
        \hline
    \end{tabular}
\end{wraptable}

For the above three tasks, we benchmark state-of-the-art models, including LLaVA-1.5~\cite{llava15}, MiniGPT-v2 \cite{chen2023minigptv2}, and GeoChat~\cite{kuckreja2024geochat}, to demonstrate the potential of LVLMs for remote sensing image understanding. LLaVA-1.5~\cite{llava15}, MiniGPT-v2 \cite{chen2023minigptv2}, Mini-Gemini~\cite{gemini}, and GeoChat~\cite{kuckreja2024geochat} are generalist models that are naturally designed for general-proposed image understanding. We, therefore, report the performance of these methods under joint training of all three tasks, i.e., image captioning, visual grounding, and VQA.

Note that all these comparison methods include a multi-stage training process. To ensure a fair comparison, we reload the models that are initially trained on large-scale image-text alignment datasets, and then finetune each method using the training set of our {\papernameAbbrev} dataset. We employ CLIP-ViT(L-14)~\cite{radford2021learning} as the vision encoder and use the Vicuna-7B model~\cite{vicuna2023} as the Large Language Model (LLM). For LLaVA-1.5~\cite{llava15}, Mini-Gemini~\cite{gemini}, and GeoChat~\cite{kuckreja2024geochat}, we adhere to the original model specifications, utilizing two MLP layers with GeLU activation~\cite{hendrycks2016gaussian}. For MiniGPT-v2\cite{chen2023minigptv2}, we implement a single MLP layer as described in the original paper. For each comparing method, we finetune the model on the training set of {\papernameAbbrev} dataset for 5 epochs. 
Following GeoChat~\cite{kuckreja2024geochat}, we use LoRA~\cite{hu2021lora} finetuning to finetune all comparing methods, with a rank of 64. 
To understand the benefit of fientuning on {\papernameAbbrev}, we include the baseline GeoChat~\cite{kuckreja2024geochat} without training on our {\papernameAbbrev} dataset for comparison. 

We further evaluate the performance of GPT-4V, which is generally known as one of the most powerful close-source vision-language models, on three tasks based on our {\papernameAbbrev} dataset. To achieve this, we directly call GPT-4V API to generate detailed captions, referring object locations, and answers for visual questions, with the following instructions. Note that we do not include object information in this experiment.

\subsection{Detailed Image Caption}
\noindent \textbf{Evaluation metrics}. For model evaluation, we follow standard practices by utilizing a set of established metrics including BLEU~\cite{papineni2002bleu}, ROUGE\_L~\cite{chin2004rouge}, METEOR~\cite{banerjee2005meteor}, and CIDEr~\cite{vedantam2015cider}. For BLEU, we consider n-gram precision with n values of 1, 2, 3, and 4. We also report average caption lengths to assess the details of generated captions. Furthermore, we note that traditional caption evaluation metrics may not be suitable for long captions. We, therefore, use an LLM-based caption evaluation metric called CHAIR\footnote{https://github.com/davidmchan/clair} proposed in \cite{chan2023clair} for our detailed image caption task. 

\noindent \textbf{Results.} Table \ref{tab_cap} shows the comparative performance of different methods in detailed image captioning of our {\papernameAbbrev} dataset. 
As demonstrated in the table, the baseline GeoChat model, when not finetuned on the {\papernameAbbrev} dataset, exhibits significantly poorer performance compared to models that have been finetuned on {\papernameAbbrev}. The LLaVA-1.5~\cite{llava15} model that has undergone fine-tuning on {\papernameAbbrev} achieves the highest performance, reaching a BLEU-1 score of 48.1 and a CIDEr score of 33.9. The GPT-4V model shows the best performance on the CHAIR score. This is expected because the CHIAR score itself is calculated using GPT-4 in our experiments. Moreover, the generated captions have an average word length of 49, which closely approximates the average length of ground truth captions. Note that detailed image captioning is a more challenging task than conventional image caption, therefore, the performance falls far behind. More advanced vision-language modeling techniques are desired to handle this challenging task.

\begin{table}[!h]
    \centering
    \resizebox{0.98\textwidth}{!}{
    \begin{tabular}{l|ccccccc|c|c}
        \toprule
        Method & BLEU-1 & BLEU-2 & BLEU-3 & BLEU-4 & METEOR & ROUGE\_L & CIDEr & CHAIR & Avg\_L \\
        \hline
        GeoChat w/o ft~\cite{kuckreja2024geochat} & 13.9 & 6.6 & 3.0 & 1.4 & 7.8 & 13.2 & 0.4 & 0.42 & 36 \\
        GPT-4V~\cite{gpt4} & 37.2 & 22.5 & 13.7 & 8.6 & 20.9 & 30.1 & 19.1 & \textbf{0.83} & 67 \\
        \midrule
        MiniGPT-v2 \cite{chen2023minigptv2} & 36.8 & 22.4 & 13.9 & 8.7 & 17.1 & 30.8 & 21.4 & 0.73 & 37 \\
        LLaVA-1.5~\cite{llava15} & \textbf{48.1} & \textbf{31.5} & \textbf{21.2} & \textbf{14.7} & \textbf{21.9} & \textbf{36.9} & \textbf{33.9} & 0.78 & 49 \\
        GeoChat~\cite{kuckreja2024geochat} & 46.7 & 30.2 & 20.1 & 13.8 & 21.1 & 35.2 & 28.2 & 0.77 & 52 \\
        Mini-Gemini~\cite{li2024mini} & {47.6} & {31.1} & {20.9} & {14.3} & {21.5} & {36.8} & {33.5} & 0.77 & 47\\
        \bottomrule
    \end{tabular}
    }
    \caption{Detailed image caption performance on {\papernameAbbrev} dataset. Avg\_L denotes the average word length of generated captions. Boldface indicates the best performance.}
    \label{tab_cap}
\end{table}

\subsection{Visual Grounding}
\noindent \textbf{Benchmark settings}. In this study, we focus on the grounded localization task, which aims to predict bounding boxes for referring objects. In our experiments, we use horizontal bounding boxes for model training and evaluating the grounding performance. Results on OBB visual grounding can be found in the supplementary. 

\noindent \textbf{Evaluation metrics}. For model evaluation, we employ the metric accuracy@$\tau$ to assess performance. Accuracy is determined by calculating the Intersection over Union (IoU) between the predicted bounding box and the ground-truth box. A prediction is considered accurate if the IoU exceeds the threshold $\tau$. In our experiments, we choose two different IoU thresholds, i.e., 0.5 and 0.7.

\noindent \textbf{Results.} Table \ref{tab_cap} shows the visual grounding performance of different methods on our {\papernameAbbrev} dataset. From the table, the model finetuned on the {\papernameAbbrev} significantly outperforms the baseline GeoChat model without finetuning. Furthermore, all models demonstrate superior performance in tasks involving unique object referring compared to non-unique object referring. This superiority is expected, as it is generally easier to localize objects uniquely identified within their categories than to differentiate among multiple instances within the same category. Note that even though MiniGPT-v2 gets worse overall grounding performance, it performs better at grounding non-unique objects.

Furthermore, GPT-4V exhibits markedly inferior performance compared to models specifically trained on image captioning and visual grounding tasks, primarily due to the absence of object information in its prompts. Despite the notable successes of existing closed-source multimodal large language models (LLMs), such as GPT-4, in comprehending natural images, their effectiveness is notably reduced when not fine-tuned on remote sensing imagery.

More importantly, even the best-performing GeoChat model fails to achieve satisfactory performance levels, with a grounding accuracy of 49.8\% at a threshold of 0.5. This shortfall is attributed to the demanding scenarios presented in the {\papernameAbbrev} dataset, which includes multiple instances of the same category as the target object. This highlights the necessity for more advanced vision grounding techniques to effectively tackle these complexities.

\begin{table}[!h]
    \centering
    \resizebox{0.98\textwidth}{!}{
    \begin{tabular}{l|cccccccc}
        \toprule
        Method & \multicolumn{2}{c}{Unique} & & \multicolumn{2}{c}{Non Unique}  & & \multicolumn{2}{c}{All} \\
        \cline{2-3}\cline{5-6}\cline{8-9}
         & Acc@0.5 & Acc@0.7 & & Acc@0.5 & Acc@0.7 & & Acc@0.5 & Acc@0.7 \\
        \hline
        GeoChat w/o ft~\cite{kuckreja2024geochat} & 20.7 & 5.4 & & 7.3 & 1.7 & & 12.9 & 3.2\\
        GPT-4V~\cite{gpt4} & 8.6 & 2.2 & & 2.5 & 0.4 & & 5.1 & 1.1 \\
        \midrule
        MiniGPT-v2 \cite{chen2023minigptv2} & 40.7 & 18.9 & & 32.4 & 15.2 & & 35.8 & 16.8 \\
        LLaVA-1.5~\cite{llava15} & 51.1 & 16.4 & & 34.8 & 11.5 &  & 41.6 & 13.6 \\
        GeoChat~\cite{kuckreja2024geochat} & \textbf{57.4} & \textbf{22.6} & & \textbf{44.5} & \textbf{18.0} & & \textbf{49.8} & \textbf{19.9} \\
        Mini-Gemini~\cite{li2024mini} & 41.1 & 9.6 &  & 22.3 & 4.9 & & 30.1 & 6.8 \\
        \bottomrule
    \end{tabular}
    }
    \caption{Visual grounding performance on {\papernameAbbrev} dataset. Boldface indicates the best performance.}
    \label{tab_ref}
\end{table}

\subsection{Visual question answering}
\noindent \textbf{Evaluation metrics}. We categorize the questions in the test set into 10 distinct types: object category, presence, quantity, color, shape, size, position, direction, scene characteristic, and reasoning. The first eight categories relate to object-level questions, whereas the last two are aligned with scene-level, and reasoning-level questions, respectively. We present the overall accuracy as well as the accuracy for each individual question type. To ensure a robust evaluation, we use GPT-4 to determine for each question whether the answers match ground truth texts, with the prompt: \textit{``Question: \{question\}, Ground Truth Answer: \{ground\_truth\}, Predicted Answer: \{predicted answer\}. Does the predicted answer match the ground truth? Answer 1 for match and 0 for not match. Use semantic meaning not exact match. Synonyms are also treated as a match, e.g., pond and swimming pool.''}.

\noindent \textbf{Results.} Table \ref{tab_qa} shows the VQA performance of different methods on our {\papernameAbbrev} dataset. As shown in the table, the baseline GeoChat~\cite{kuckreja2024geochat} model without finetuning gets an average accuracy of 40.8\%. Further finetuning on our {\papernameAbbrev} training set significantly boosts the average accuracy to 76.0\%. GPT-4V gets a reasonable performance but still falls a lot behind fine-tuned models, suggesting that detailed object information contributes a lot to the visual question answering task on our benchmark.

\begin{table}[!h]
    \centering
    \resizebox{0.98\textwidth}{!}{
    \begin{tabular}{l|cccccccccc|c}
        \toprule
        Method & Category & Presence & Quantity & Color & Shape & Size & Position & Direction & Scene & Reasoning & All \\
        \hline
        \# VQAs & 5435 & 7789 & 6374 & 3550 & 1422 & 1011 & 5829 & 477 & 4620 & 902 &  \\
        \hline
        GeoChat w/o ft~\cite{kuckreja2024geochat} & 48.5 & 85.9 & 19.2 & 17.0 & 18.3 & 32.0 & 43.4 & 42.1 & 44.2 & 57.4 & 40.8\\
        GPT-4V~\cite{gpt4} & 67.0 & 87.6 & 45.6 & 71.0 & 70.8 & 54.3 & 67.2 & 50.7 & 69.8 & 72.4 & 65.6 \\
        \midrule
        MiniGPT-v2 \cite{chen2023minigptv2} & 61.3 & 26.0 & 46.1 & 51.0 & 41.8 & 11.2 & 17.1 & 12.4 & 49.3 & 21.9 & 38.2 \\
        LLaVA-1.5~\cite{llava15} & 86.9 & 91.8 & 58.2 & 69.9 & 72.2 & \textbf{61.5} & \textbf{69.5} & \textbf{56.7} & \textbf{83.9} & 73.4 & 76.4 \\
        GeoChat~\cite{kuckreja2024geochat} & 86.5 & \textbf{92.1} & 56.3 & 70.1 & 73.8 & 60.4 & 69.3 & 53.5 & 83.7 & 73.5 & 76.0  \\
        Mini-Gemini~\cite{li2024mini} & \textbf{87.8} & 92.1 & \textbf{58.8} & \textbf{74.0} & \textbf{75.3} & 58.0 & 68.0 & 56.7 & 83.2 & \textbf{74.4} & \textbf{77.8} \\
        
        \bottomrule
    \end{tabular}}
    \caption{Visual question answering performance on {\papernameAbbrev} dataset. Boldface indicates the best performance.}
    \label{tab_qa}
\end{table}

\section{Related Work}
\subsection{Remote Sensing Image Captioning Datasets}
Image captioning in remote sensing is a well-established task that focuses on creating descriptive text for overhead imagery. Commonly used datasets such as UCM-Captions~\cite{qu2016deep}, Syndey-Captions~\cite{qu2016deep}, and RSICD \cite{lu2017exploring} have been instrumental by offering brief scene descriptions. However, these datasets typically provide short and less detailed captions that overlook intricate object details. Recent efforts, such as RSGPT~\cite{hu2023rsgpt}, have introduced high-quality, human-generated detailed captions, though the dataset is limited to just 2,585 image-text pairs, which hampers its utility for developing robust vision-language models in remote sensing. In contrast, RS5M~\cite{zhang2023rs5m} introduced a substantial dataset featuring 5 million detailed captions. However, these captions are generated automatically, resulting in quality that is not guaranteed. In stark contrast, our {\papernameAbbrev} dataset includes \Nimages human-verified captions that are not only of high quality but also rich in detail, ensuring both reliability and practical utility for advanced remote sensing applications.

\subsection{Remote Sensing Visual Grounding Datasets}
Visual grounding in remote sensing has recently emerged as an intriguing field of study. Unlike referring expressions in natural images, those in RSVG frequently involve complex geospatial relationships, and the objects of interest may not be prominently visible. The first RSVG dataset was introduced in~\cite{sun2022visual}, featuring 4,239 images from GoogleEarth and 7,993 referring expressions. Subsequently, Zhan et al.~\cite{zhan2023rsvg} introduced the DIOR-RSVG dataset, which includes 17,402 remote sensing images and 38,320 referring expressions across 20 object categories. Recent studies~\cite{yuan2024rrsis,liu2024rotated} have developed visual grounding datasets for remote sensing that include object segmentation; however, these tend to be smaller in scale. In contrast, our {\papernameAbbrev} dataset incorporates a substantial number of object-referring expressions.

\subsection{Remote Sensing Visual Question Answering Datasets}
RSVQA~\cite{lobry2020rsvqa} established the first VQA benchmark dataset for remote sensing images. This dataset comprises RS images sourced from OpenStreetMap, accompanied by automatically generated questions and answers. It includes 772 images with 77,232 question-answer pairs in the low-resolution collection and 10,659 images with 1,066,316 pairs in the high-resolution collection. 
Zheng et al.\cite{zheng2021mutual} launched the RSIVQA dataset, a remote sensing VQA dataset that features approximately 37k images and 110,000 question-answer pairs. A small portion of question-answer pairs in RSIVQA are annotated by human annotators.
Al et al.\cite{al2022open} introduced an innovative dataset, VQA-TextRS, which consists of 2,144 RS images and 6,245 question-answer pairs generated and annotated by humans in an open-ended format. More recently, the RSIEval\cite{hu2023rsgpt} dataset features 936 human-crafted question-answer pairs from 100 remote sensing images. Similarly, our {\papernameAbbrev} dataset also incorporates open-ended question-answer pairs, created by GPT-4V and validated by human annotators, with \Nvqa question-answer pairs.

\section{Conclusion and future work}
In this work, we introduce {\papernameAbbrev}, a versatile vision-language dataset and benchmark for remote sensing image understanding. This comprehensive dataset not only addresses the limitations of previous datasets that either ignore detailed object information or suffer from quality control issues but also enriches the field by providing a diverse range of annotations including detailed captions, object referring, and visual question answering with rich object information and verified by human annotators. Our benchmark challenges, specifically designed around the {\papernameAbbrev} dataset, demonstrate the practical utility of our dataset in advancing the capabilities of vision-language models in the domain of remote sensing.

Currently, the {\papernameAbbrev} dataset is limited to annotations for RGB images. In future work, we aim to enhance {\papernameAbbrev} by incorporating annotations from a variety of remote sensing data types, including infrared images, multi- and hyperspectral images, Synthetic Aperture Radar (SAR) images, and temporal datasets. This expansion will significantly broaden the dataset’s utility across diverse observation conditions, facilitating more accurate and timely applications in remote sensing.

\section{Broader Impact}
{\papernameAbbrev} provides a comprehensive benchmark for developing and evaluating generalist vision-language models in both remote sensing and computer vision. This dataset not only supports the training and evaluation of advanced vision-language models but also boosts their ability to tackle complex real-world scenarios in remote sensing.

\newpage
{
\small
\bibliographystyle{unsrt}
\bibliography{egbib} 
}

\appendix

\newpage

\section{{\papernameAbbrev} Documentation and Intended Uses}

\subsection{Overview}
{\papernameAbbrev} consists of \Nimages remote sensing images with detailed captions, \Nref object refers, \Nvqa visual question-answer pairs. 
{\papernameAbbrev} is designed to facilitate the development and evaluation of vision-language models in remote sensing, providing a comprehensive set of annotations including detailed captions, visual grounding, and visual question answering. This section documents the dataset in accordance with best practices to ensure transparency, reproducibility, and ethical usage. 

\subsection{Data Organizing}
Our {\papernameAbbrev} dataset is organized as follows.

\dirtree{%
.1 root/.
.2 Images\_train.zip.
.2 Annotation\_train.zip.
.2 Images\_val.zip.
.2 Annotation\_val.zip.
.2 {\papernameAbbrev}\_train.json.
.2 {\papernameAbbrev}\_EVAL\_Cap.json.
.2 {\papernameAbbrev}\_EVAL\_referring.json.
.2 {\papernameAbbrev}\_EVAL\_vqa.json.
}

Detailed descriptions for each folder or file are given below.
\begin{itemize}
    \item Images\_train.zip contains all raw images in the training split.
    \item Annotation\_train.zip contains all annotations in the training split, one JSON file per image.
    \item Images\_val.zip contains all raw images in the validation split.
    \item Annotation\_val.zip contains all annotations in the validation split, one JSON file per image.
    \item {\papernameAbbrev}\_train.json contains all training annotations following LLaVA in standard JSON format.
    \item {\papernameAbbrev}\_EVAL\_Cap.json contains all evaluation annotations for the captioning task in standard JSON format.
    \item {\papernameAbbrev}\_EVAL\_referring.json contains all evaluation annotations for the visual grounding task in standard JSON format.
    \item {\papernameAbbrev}\_EVAL\_vqa.json contains all evaluation annotations for the VQA task in standard JSON format. 
\end{itemize}

\subsection{Intended Uses}

{\papernameAbbrev} is intended for use in academic and research settings, specifically for:
\begin{itemize}
    \item Training and evaluating vision-language models capable of understanding complex visual and textual tasks.
    \item Advancing the state-of-the-art in remote sensing image analysis by providing a rich dataset that supports multiple tasks.
\end{itemize}

\subsection{Use Cases}

\begin{itemize}
    \item \textbf{Academic Research}: {\papernameAbbrev} is ideal for exploring new algorithms in image captioning, visual grounding, and visual question answering within the remote sensing domain.
    \item \textbf{Model Evaluation}: The dataset can serve as a benchmark for comparing different vision-language models’ performance on a standardized set of tasks.
    \item \textbf{Educational Purposes}: The dataset and its comprehensive annotations can be used in coursework and workshops to teach advanced techniques in machine learning and remote sensing.
\end{itemize}

\subsection{Limitations}

\begin{itemize}
    \item \textbf{Geographic Diversity}: While {\papernameAbbrev} includes a variety of landscapes, the geographic diversity is limited to the regions covered by the DOTA-v2 and DIOR datasets.
    \item \textbf{Annotation Bias}: Despite efforts to ensure high-quality annotations through human verification, biases may exist in the interpretations of visual data due to subjective human factors.
\end{itemize}

\subsection{Ethical Considerations}

\begin{itemize}
    \item \textbf{Privacy and Sensitivity}: The dataset consists of non-sensitive, publicly available satellite images where no individual person or private property can be identified.
    \item \textbf{Use Restrictions}: Users are encouraged to use {\papernameAbbrev} responsibly and ethically, particularly when developing applications that might impact environmental monitoring and urban planning.
\end{itemize}

\subsection{Documentation and Maintenance}

\begin{itemize}
    \item \textbf{Versioning}: Detailed version history of the dataset will be maintained to track changes and improvements over time.
    \item \textbf{Community Involvement}: Feedback from the user community is encouraged to improve the dataset’s quality and applicability in various use cases.
\end{itemize}

\subsection{Statements for NLP} 
We employ GPT-4V~\cite{gpt4} to generate initial annotations; for further details, please refer to the main paper. These annotations undergo a manual review by human annotators.

\subsection{Accountability Framework}
To ensure responsible usage and continuous improvement, an accountability framework is established. Users of {\papernameAbbrev} are encouraged to report any issues or biases they encounter, contributing to an ongoing effort to refine the dataset and its annotations.

\section{Dataset Collection Details}

\begin{itemize}
    \item \textbf{Source datasets}: Images are sourced from the DOTA-v2~\cite{ding2021object} and DIOR~\cite{li2020object} datasets and annotated with high-resolution details. 
    We divide each image into patches measuring 512 × 512 pixels and filter out patches with no object annotations. This yields over 20,310 image patches from the DOTA-v2 dataset and 9,304 patches from the DIOR dataset. Statistics are given in Table~\ref{tab_det_data}.
    \item \textbf{Preprocessing}: We extract image-level information and object-level information for all image patches. Note that the original DOTA-v2 and DIOR datasets contain 18 and 20 object categories respectively. We merge shared object categories and also merge small-vehicle and large-vehicle into the vehicle category. After merging, we get 26 object categories, including airplane, airport, baseball-diamond, basketball-court, bridge, chimney, container-crane, dam, expressway-service-area, expressway-toll-station, golf-field, ground-track-field, harbor, helicopter, helipad, overpass, roundabout, ship, soccer-ball-field, stadium, storage-tank, swimming-pool, tennis-court, train-station, vehicle, windmill.
    \item \textbf{Object attribute extraction}: 
    We then extract object attributes and formulate a JSON file for each image patch, including object category, corner points, bounding box, position, relative position, size, and relative size. We also determine whether each object is unique within its category or not. We do not extract object colors because objects can have complex structures with multiple colors, and we rely on GPT-4V to identify object colors. The code for preprocessing is provided at \url{\gitpage}.
    \item \textbf{GPT-4V annotation generation}: The code for GPT-4V inference is provided at \url{\gitpage}. Detailed instructions are provided in Section \ref{sc_prompt}.
\end{itemize}

\begin{table}[!h]
    \centering
    \begin{tabular}{l|cccc}
        \toprule
        Dataset & \#Images & \#Valid Patches & \#Selected Patches & Category \\ 
        \hline
        DOTA-v2~\cite{ding2021object} & 2,423 & 29,910 & 20,310 & 18 \\
        DIOR~\cite{li2020object} & 11,725 & 9,304 & 9,304 & 20 \\
        \bottomrule
    \end{tabular}
    \caption{Statistics of source object detection datasets.}
    \label{tab_det_data}
\end{table}

\section{URL to Data and Metadata}
The {\papernameAbbrev} dataset can be accessed and downloaded through our dedicated platform, which provides detailed views of the dataset components and their annotations.

For practical examples and to download the dataset, visit our Huggingface repository (\url{\huggingpage}). Detailed metadata for the dataset is documented using the Croissant metadata framework, ensuring comprehensive coverage and compliance with the MLCommons Croissant standards, check [metadata](\url{https://huggingface.co/api/datasets/xiang709/VRSBench}). Please check our Huggingface repo for metadata details.

\section{Author Statement and Data License}

\textbf{Author Responsibility Statement:} The authors bear all responsibilities in case of any violations of rights or ethical concerns regarding the {\papernameAbbrev} dataset. 

\textbf{Data License Confirmation:} The dataset is released under the [CC-BY-4.0], which permits unrestricted use, distribution, and reproduction in any medium, provided the original work is properly cited.

\section{Hosting and Accessibility} The {\papernameAbbrev} dataset is hosted on Huggingface (\url{\huggingpage}) to ensure reliable and continuous accessibility. 

\textbf{Maintenance Plan:} Ongoing maintenance and updates will be managed by the dataset authors, with updates scheduled bi-annually or as significant changes in the data sources occur.

\textbf{Long-term Preservation:} The dataset is archived in Huggingface (\url{\huggingpage}) to ensure long-term availability.

\textbf{Structured Metadata:} The annotation for each image is well-organized in standard JSON format to ensure easy usage.


\section{Data Creation Details}

\subsection{GPT-4V Prompts}\label{sc_prompt}
We carefully design the following instructions to prompt GPT-4V to generate annotations of image captions, referring sentences, and visual question-answering pairs. 

``You are an AI visual assistant tasked with analyzing remote sensing images. For each image, you receive image meta information and a list of objects in the format: \{image source: image source, image resolution: image resolution, objects: [{obj\_id: object id, obj\_cls: object category, obj\_corner: corner point, obj\_coord: object bounding box, is\_unique: unique object or not, obj\_position: object position, obj\_rel\_position: object relative position within category, obj\_size: object size, obj\_rel\_size: object relative size within category, flag: refer or not}, ...]\}. The bounding box coordinates [x1, y1, x2, y2] are floating numbers from 0 to 1, corresponding to the top left x, top left y, bottom right x, and bottom right y. Note that the top-left corner coordinates are (0,0) and the bottom-right corner coordinates are (1,1).

Your job is to create a detailed image caption and referring sentences for 1-5 distinct objects, if multiple are present, as well as a list of question-answer pairs. Each referring sentence should unambiguously refer to one object. Finally, you need to return a JSON file in the format: \{caption: detailed image caption, objects: [{obj\_id: object id, ref: referring sentence},...], qa\_pairs: [ques\_id: question id, question: question, type: question type, answer: answer]\}. Do not return any notes after the JSON.

Here are further important instructions for referring sentences: \\
1. Identify 1-5 distinguishable objects and provide referring sentences. Each sentence alone must independently, without seeing others, and unambiguously identify an object.\\
2. Select all unique objects (is\_unique=True) for creating referring sentences. Do not select objects whose flag=True for referring sentences, but still use them for captioning and question-answering tasks.\\
3. Use distinctive features to describe objects. Try to use diverse object attributes such as color, shape, position, size, relative position, and relative size, but avoid specifying size details for small or large vehicles. Some object attributes are not provided, you may need to identify them from the input image. Do not explain why it is distinctive or distinguishable.\\
4. For each object category, select only 1-3 most distinguishable objects and ensure the referring sentences can confidently distinguish each of them from other objects of the same category.\\
5. Avoid ordinal descriptors and references (first-mentioned, aforementioned, or previously mentioned) to prior mentions. Instead, use distinct features to refer back to previously identified objects.\\
6. If multiple object categories exist, try to include diverse object categories in a balanced manner.\\
7. For referring sentences, use natural language to describe objects based on their bounding box data, without directly mentioning the coordinates. Do not mention whether the object is distinguishable or not.\\
8. You may include roads/bridges running east-west or north-south but do not mention object-facing directions or pointing directions.\\
9. Do not mention the noses, vertical stabilizers, tails, or tail fins of planes, airplanes, or aircraft \\
10. Do not mention gate numbers when describing airports or airplanes.\\
11. Carefully verify each piece of information before finalizing the referring sentences, make sure each referring sentence alone can distinguish one object without any ambiguity. If not, remove this referring object.

Here are further important instructions for image captioning: \\
1. Create a detailed caption for the provided image, incorporating all visible elements and object information. Focus on describing the content of the image without mentioning the reference status of objects or their flag status.\\
2. Start the caption with an overview of the image. Possibly include the source of the image (if provided), specify whether it is in grayscale or color, and mention the resolution (if provided). Follow this with a description of specific, clear details within the image. Summarize the image's content in 3-7 sentences, making sure to include counts of prominent objects.\\
3. Describe only clear features; avoid uncertainties and unclear elements. Do not mention anything that is unknown or not specified.\\
4. Possibly include other visual objects in the image that are not provided as inputs, such as buildings, houses, roads, and trees if they are obvious and non-ambiguous.\\
5. Highlight diverse object attributes such as color, shape, position, size, relative position, and relative size. Do not add size details for small or large vehicles. \\
6. Exclude imagined details not visible in the image, like weather, people, or object usage. Do not imagine the moving status of airplanes, ships, or vehicles if you are not sure about it.\\
7. For roads, include features like shape (straight/curved), width, length, and orientation.\\
8. For houses, mention characteristics like density, size, rooftop color, and presence of gardens.\\
9. For airports, include details like boarding bridges, terminals, boarding ports, and tarmac.\\
10. Carefully verify each piece of information before finalizing the caption.\\
11. Do not mention whether the image is taken during the day or night.\\
12. Do not mention whether the vehicles are in motion or not.

Here are questions for visual question answering: \\
1. Based on all visible elements and object information, ask 3-10 questions about diverse types, including object category, object existence, object quantity, object color, object shape, object size, object position, object direction, scene type, rural or urban, and reasoning. The category of scene type includes the main structure/type of area. Additionally, the category of reasoning is available for questions that require multifaceted analytical thought processes (e.g., object distribution pattern). Possibly include objects that are not provided, such as houses, roads, and trees if they are obvious and non-ambiguous. \\
2. Do not mention the object referred or not. Do not mention any flag information in questions and answers.\\
3. Ensure each question has a definite answer without any ambiguity, and answer each question using a single word or phrase, no more than 3 words.\\
4. When answering questions about the number of objects, take into account all object information.\\
5. Only ask questions about clear answers; avoid uncertainties or unclear elements, such as unknown, uncertain, some, or several. If the answer is uncertain or unable to be determined, remove this question-answer pair.\\
6. Do not use first, second, third, fourth, fifth, first-mentioned, or previously mentioned to refer to objects, use distinguishable features to refer to specific objects mentioned before.\\
7. Try to cover diverse types of questions.\\
8. Do not ask about the type of view the image was captured from, or whether the image was taken during day or night. \\
9. Do not ask the source of the image. \\
10. Do not ask facing direction, but you may ask whether roads/bridges running east-west or north-south. \\
11. Do not ask whether the vehicles are in motion or not. \\
12. Do not ask whether the image is taken during the day or night''.

\subsection{Human verification guidelines}

Given an input image and associating detailed image caption, check if each piece of provided information is correct or not (no check for image source). If incorrect, correct the information, possible corrections include modifying/removing words/sentences. Modification is preferred to removing. But if a caption sentence, referring sentence, or question-answer pair is totally wrong/ambiguous/uncertain, remove it. 
\begin{itemize}
    \item For caption annotations: Make sure each piece of information in the caption is correct. Remove uncertain or meaningless elements. Be careful of object counts, take into account all objects, both referred and not referred.
    \item For object referring annotations: Make sure each referring sentence can distinguishably identify the correct object (numbered in boxes) without any ambiguity. Be careful of object color/orientation.
    \item For VQA annotations: Make sure each question has a clear answer without any ambiguity, and each answer should be correct using a single word or phrase, no more than 3 words. Correct answers that specify objects by object IDs. Remove self-answered question-answer pairs.  
    \item Include question type for each QA, all possible question types include: object category, object existence, object quantity, object color, object shape, object size, object position, object direction, scene type, and reasoning. The category of scene type includes color or grayscale, main structure/type of area, and rural or urban. Additionally, the category of reasoning is available for questions that require multifaceted analytical thought processes (e.g., object distribution pattern).
\end{itemize}

\section{Experimental details}
\subsection{Training details}
In our experimental setup, all comparative methods are trained on a single node equipped with 4 Nvidia A100 GPUs. The batch size is standardized at 32, and each model undergoes training for a duration of five epochs. We initialize the learning rate at 2e-4 and employ a cosine learning rate decay schedule for optimization. The learning rate experiences a warm-up phase, reaching 3\% of the total training steps to gradually adapt to the training regime.

\subsection{Visual grounding using OBBs}
\textbf{Settings}. In the main paper, horizontal bounding boxes are utilized for both training the model and evaluating its visual grounding capabilities. This section extends the evaluation to incorporate oriented bounding boxes for object localization. Given that GeoChat has demonstrated superior performance in object grounding using bounding boxes, this experiment is exclusively dedicated to exploring the effectiveness of GeoChat under the conditions of oriented bounding boxes. In our experiments, the oriented bounding box is defined by the parameters \([cx, cy, w, h, \theta]\), where \( (cx, cy) \) represents the center coordinates, \( w \) and \( h \) represent the width and height of the bounding box, respectively, and \( \theta \) indicates the rotation angle. 

\textbf{Results}. As shown in Table~\ref{tab_ref_obb}, the GeoChat model achieves an overall grounding accuracy of 24.3\% at a threshold of 0.5, which is lower than its counterpart using horizontal bounding boxes, where the grounding accuracy reaches 49.8\%. This highlights that visual grounding with oriented bounding boxes presents a greater challenge compared to grounding with horizontal bounding boxes.

\begin{table}[!h]
    \centering
    \resizebox{0.98\textwidth}{!}{
    \begin{tabular}{l|ccccccccc}
        \toprule
        Method & \multicolumn{2}{c}{Unique} & & \multicolumn{2}{c}{Non Unique}  & & \multicolumn{3}{c}{All} \\
        \cline{2-3}\cline{5-6}\cline{8-10}
         & Acc@0.5 & Acc@0.7 & & Acc@0.5 & Acc@0.7 & & Acc@0.5 & Acc@0.7 & Err\_R\\
        \hline
        GeoChat~\cite{kuckreja2024geochat} &  32.3 & 12.6 & &  18.5 & 5.7 & & 24.3 & 8.6 & 18.7 \\
        \bottomrule
    \end{tabular}
    }
    \caption{Visual grounding performance on {\papernameAbbrev} dataset using orientated bounding boxes for referring object localization. Err\_R denotes average angle prediction error.}
    \label{tab_ref_obb}
\end{table}

\subsection{Qualitative results}
\begin{figure}[!h]
    \centering
    \begin{subfigure}[t]{\textwidth}
        \caption{}
        \includegraphics[width=\textwidth]{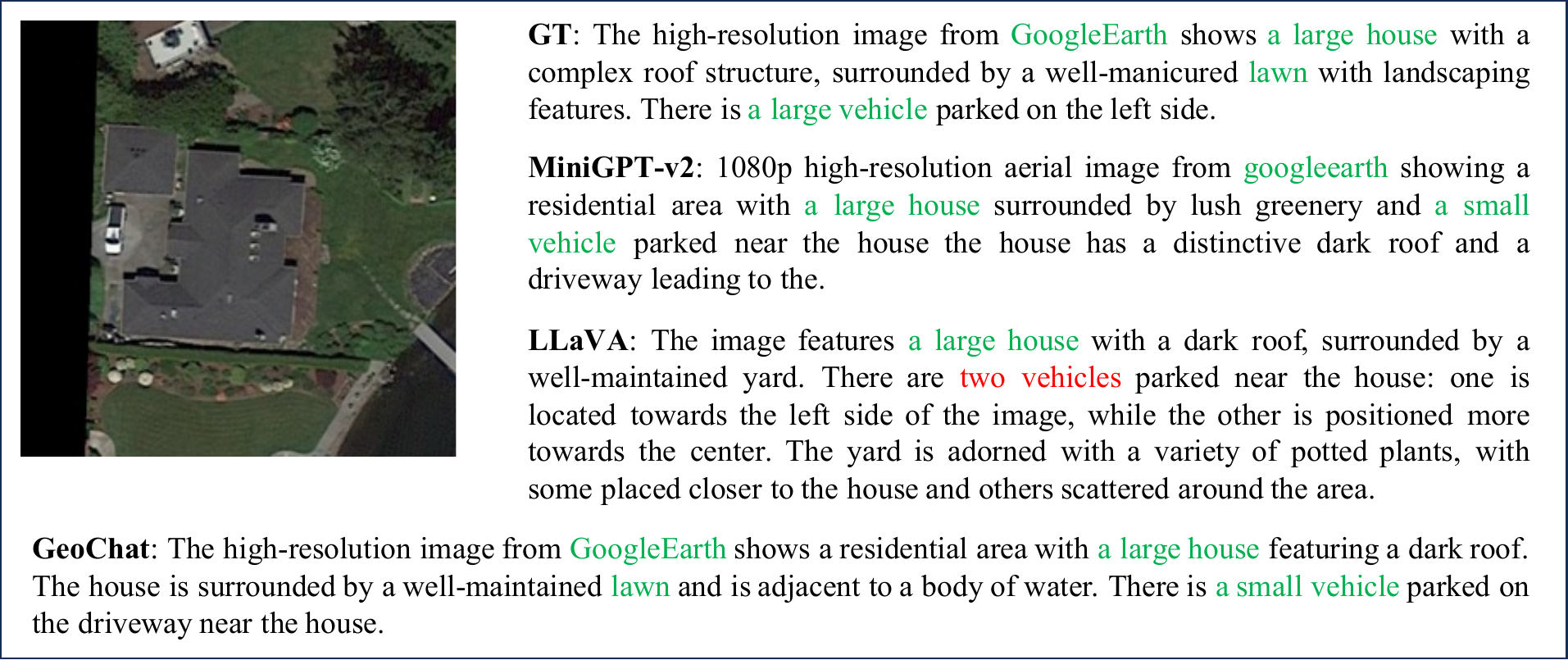}
    \end{subfigure}
    \vfill
    \begin{subfigure}[t]{\textwidth}
        \caption{}
        \includegraphics[width=\textwidth]{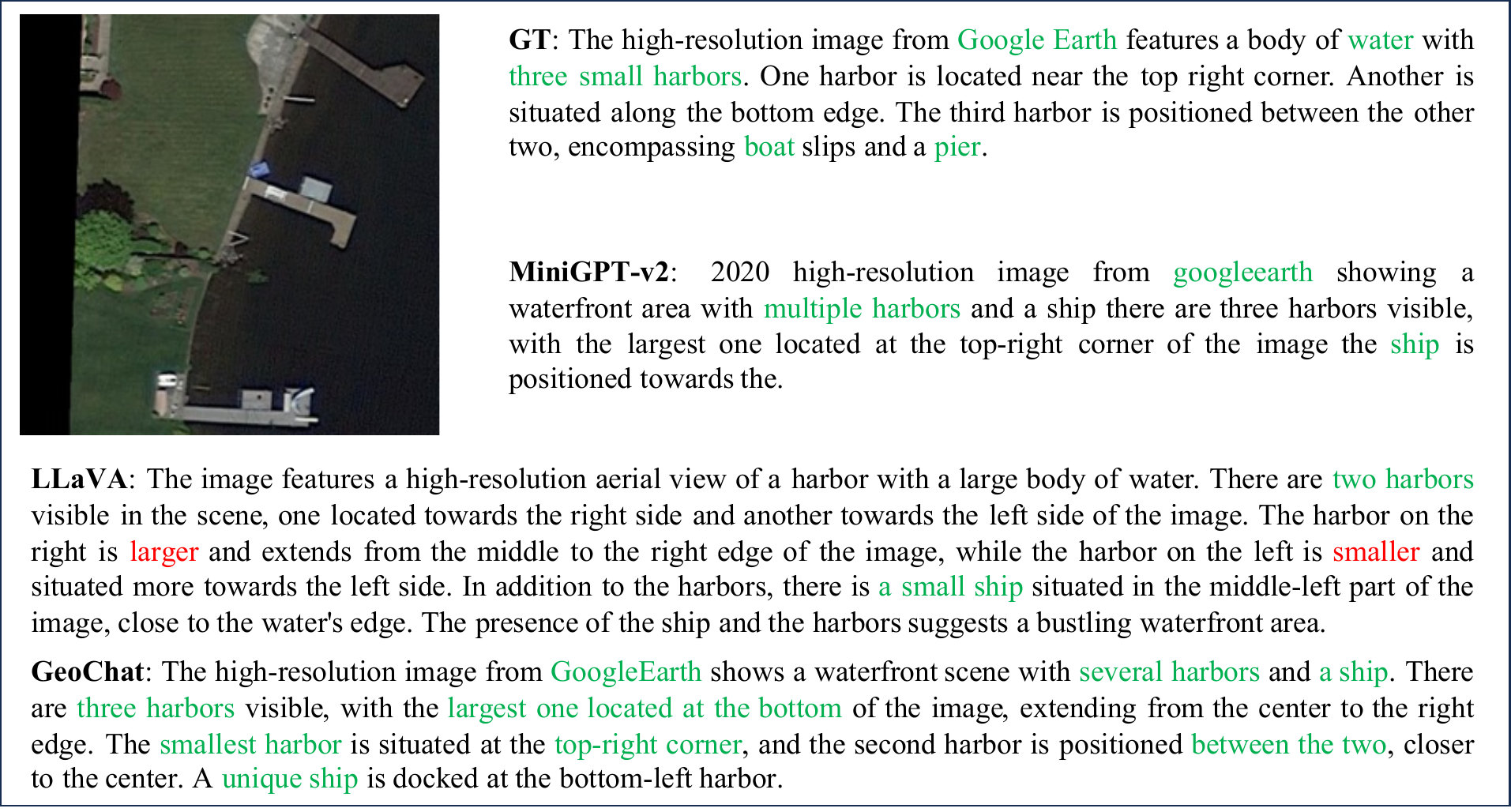}
    \end{subfigure}
    \caption{Selected examples of detailed image caption results. We highlight correct information in green and incorrect information in red.}
    \label{fig_rst_cap}
\end{figure}

\begin{figure}[!h]
    \centering
    \begin{subfigure}[t]{0.24\textwidth}
        \caption{}
        \includegraphics[width=\textwidth]{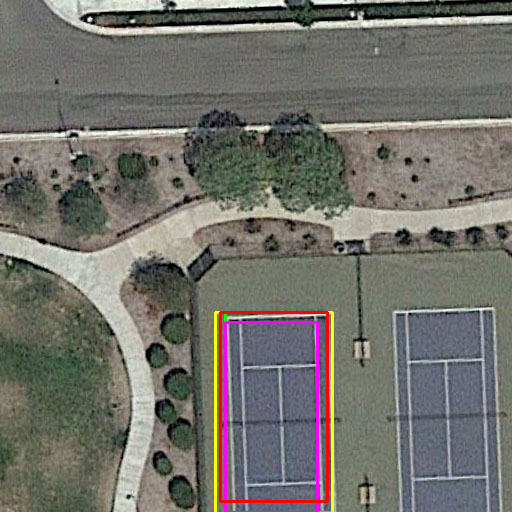}
    \end{subfigure}
    \hfill
    \begin{subfigure}[t]{0.24\textwidth}
        \caption{}
        \includegraphics[width=\textwidth]{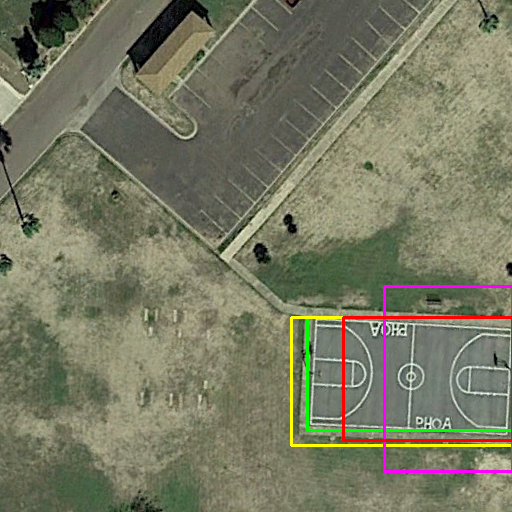}
    \end{subfigure}
    \hfill
    \begin{subfigure}[t]{0.24\textwidth}
        \caption{}
        \includegraphics[width=\textwidth]{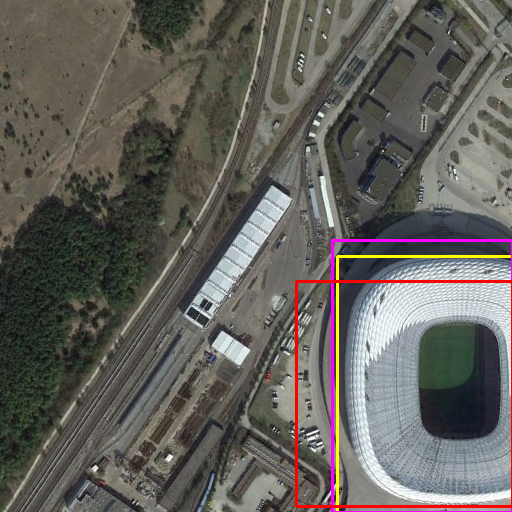}
    \end{subfigure}
    \hfill
    \begin{subfigure}[t]{0.24\textwidth}
        \caption{}
        \includegraphics[width=\textwidth]{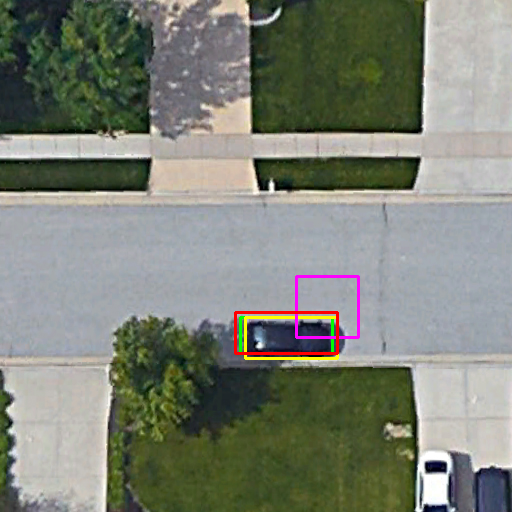}
    \end{subfigure}

    \begin{subfigure}[t]{0.24\textwidth}
        \caption{}
        \includegraphics[width=\textwidth]{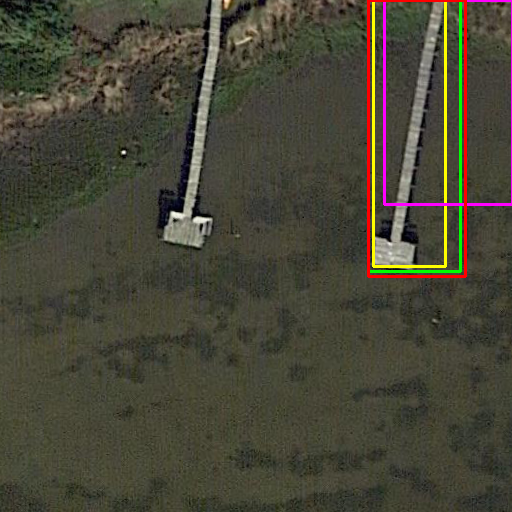}
    \end{subfigure}
    \hfill
    \begin{subfigure}[t]{0.24\textwidth}
        \caption{}
        \includegraphics[width=\textwidth]{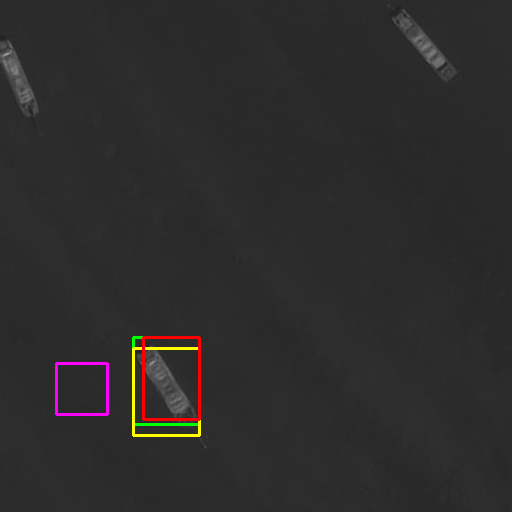}
    \end{subfigure}
    \hfill
    \begin{subfigure}[t]{0.24\textwidth}
        \caption{}
        \includegraphics[width=\textwidth]{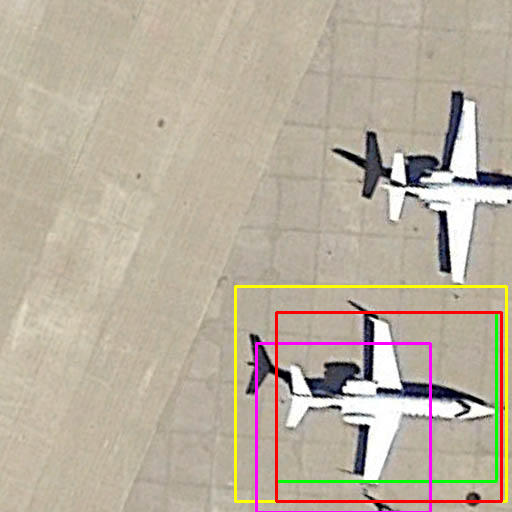}
    \end{subfigure}
    \hfill
    \begin{subfigure}[t]{0.24\textwidth}
        \caption{}
        \includegraphics[width=\textwidth]{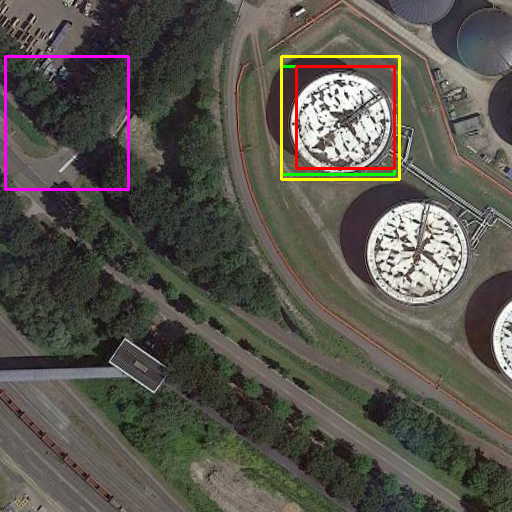}
    \end{subfigure}
    
    \begin{subfigure}[t]{0.5\textwidth}
        \includegraphics[width=\textwidth]{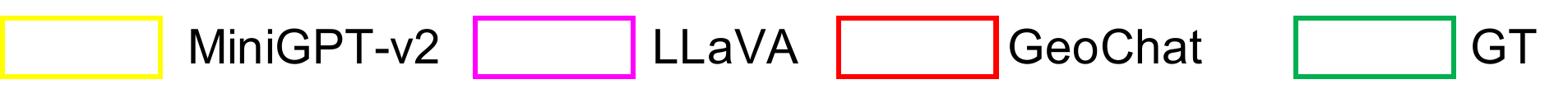}
    \end{subfigure}
    
    \caption{Selected examples of Visual grounding. (a) The tennis court on the left side of the image, surrounded by a brownish surface. (b) The basketball court located at the right side of the image. (c) The dome stadium situated towards the bottom-right side of the image. (d) The vehicle parked closest to the top edge of the image. (e) The harbor located at the right-most edge of the image. (f) The small ship located towards the bottom-left of the image. (g) The airplane is located towards the bottom of the frame. (h) The left-most storage tank is fully visible and situated on the upper side of the image.}
    \label{fig_rst_ref}
\end{figure}

\begin{figure}[!h]
    \centering
    \begin{subfigure}[t]{\textwidth}
        \caption{}
        \includegraphics[width=\textwidth]{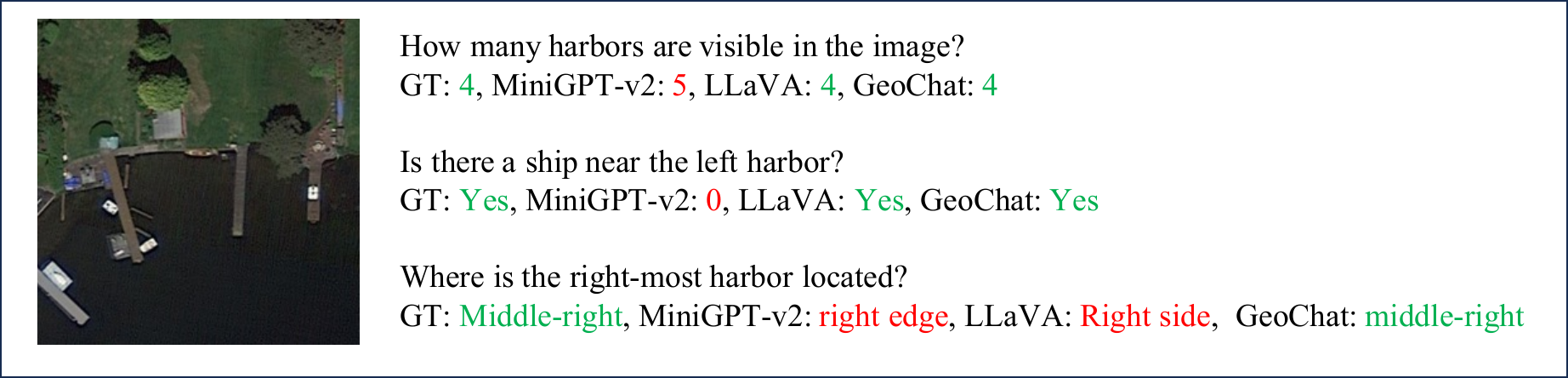}
    \end{subfigure}
    \vfill
    \begin{subfigure}[t]{\textwidth}
        \caption{}
        \includegraphics[width=\textwidth]{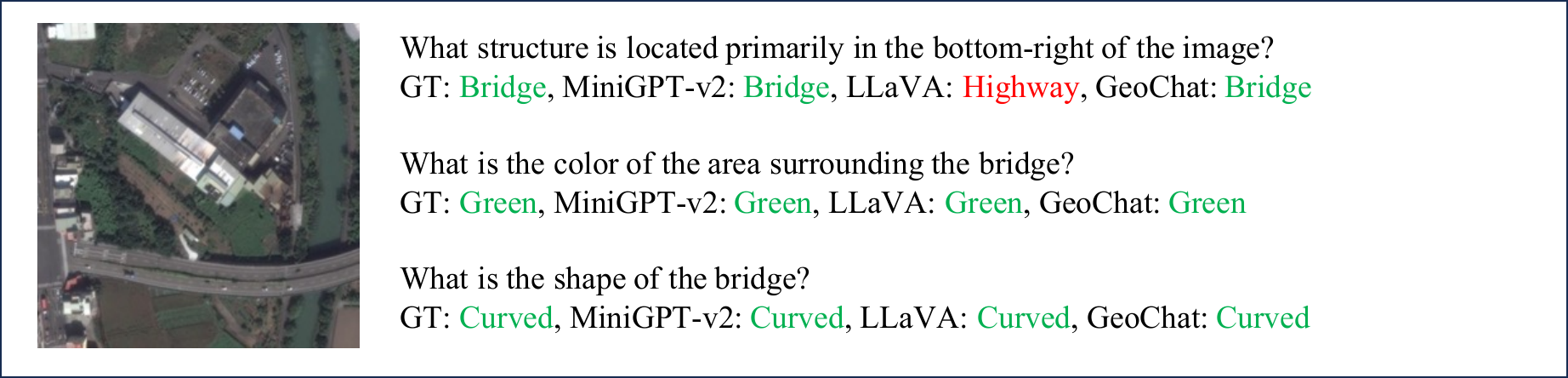}
    \end{subfigure}
    \caption{Selected examples of VQA results. Correct answers are shown in green and incorrect answers are shown in red.}
    \label{fig_rst_vqa}
\end{figure}

We show qualitative results of detailed image caption in Fig.~\ref{fig_rst_cap}, visual grounding in Fig.~\ref{fig_rst_ref}, and visual question answering in grounding in Fig.~\ref{fig_rst_vqa}.

\subsection{GPT-4V Evaluation Prompt}

"You are an AI visual assistant tasked with analyzing remote sensing images. You receive an image and multiple object-referring sentences and visual questions. Your role is to provide a detailed caption for the image, identify object locations for all referring sentences, and answer all visual questions.

Here are detailed instructions for image caption: Describe the image in detail in 3-7 sentences, making sure to include counts of prominent objects. Describe only clear features; avoid uncertainties and unclear elements. Do not mention anything that is unknown or not specified. Highlight diverse object attributes such as color, shape, position, size, relative position, and relative size. Do not add size details for small or large vehicles. Exclude imagined details not visible in the image, like weather, people, or object usage. Do not imagine the moving status of airplanes, ships, or vehicles if you are not sure about it. For roads, include features like shape (straight/curved), width, length, and orientation. For houses, mention characteristics like density, size, rooftop color, and presence of gardens. For airports, include details like boarding bridges, terminals, boarding ports, and tarmac. Do not mention the image is taken during the day or night. Do not mention whether the vehicles are in motion or not. Carefully verify each piece of information before finalizing the caption.

For each referring sentence, tell me the location of the referred object in the image, and return its bounding box coordinates in the format of [x1, y1, x2, y2], which denotes the top left x, top left y, bottom right x, and bottom right y.

Here are detailed instructions for visual question answering: For each question, answer the question based on the image content in a single word or a short phrase.

Finally, you need to return \{caption: detailed image caption, objects: [{obj\_id, referring\_sentence, location},...], qa\_pairs: [ques\_id: question id, type: question type, question: question, answer: answer]\} in JSON format. Do not return any notes after the JSON."

\end{document}